\documentclass[10pt,twocolumn,letterpaper]{article}

\usepackage{cvpr}              
\makeatletter
\@namedef{ver@everyshi.sty}{}
\makeatother
\usepackage{tikz}

\usepackage{graphicx}
\usepackage{amsmath}
\usepackage{amssymb}
\usepackage{booktabs}
\usepackage{eucal}
\usepackage{hyperref}

\usepackage{hyperref}
\hypersetup{
    colorlinks=true,
    linkcolor=blue,
    filecolor=magenta,      
    urlcolor=cyan,
    pdftitle={Overleaf Example},
    pdfpagemode=FullScreen,
    }
\newcommand{\xuyi}[1]{{\textcolor{black}{#1}}}

\newcommand{\zhongli}[1]{{\textcolor{black}{#1}}}
\usepackage{changes}

\newcommand\given[1][]{\:#1\vert\:}
\newcommand\norm[1]{\left\lVert#1\right\rVert}
\usepackage[capitalize]{cleveref}
\crefname{section}{Sec.}{Secs.}
\Crefname{section}{Section}{Sections}
\Crefname{table}{Table}{Tables}
\crefname{table}{Tab.}{Tabs.}

\newcommand{\figurePath}{figures}

\def\cvprPaperID{3527} 
\def\confName{CVPR}
\def\confYear{2022}

\begin{document}

\title{NeuLF: Efficient Novel View Synthesis with Neural 4D Light Field}

\author{Zhong Li$^1$ \qquad Liangchen Song$^2$ \qquad Celong Liu$^1$ \qquad Junsong Yuan$^2$ \qquad Yi Xu$^1$  \\$^1$OPPO US Research Center, Palo Alto, California, USA \\$^2$University at Buffalo, State University of New York, Buffalo, New York, USA\\\\ \href{<url>}{https://lizhong3232.github.io/neulf/}}



\twocolumn[{%
\renewcommand\twocolumn[1][]{#1}%
\maketitle

\begin{center}
    \centering
    \captionsetup{type=figure}
    \includegraphics[width=\textwidth]{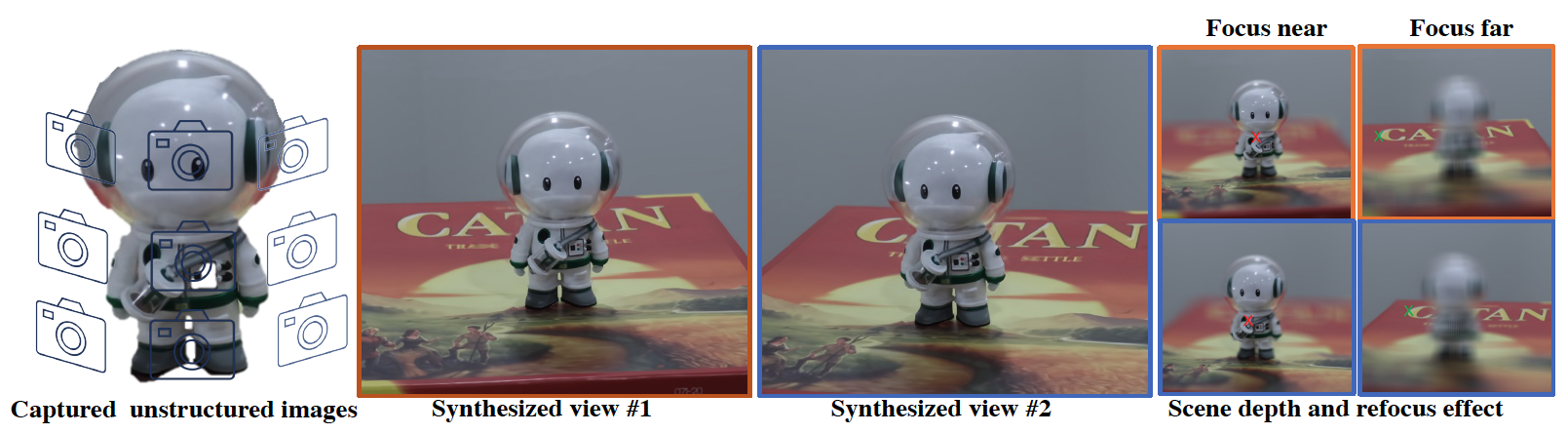}
    \captionof{figure}{Given a set of images captured in front of a scene using a handheld camera, \textbf{Neural 4D Light Field (NeuLF)} uses an implicit neural representation to learn the mapping from rays to color values. With the learned model, novel views can be synthesized by predicting the color of each ray. Moreover, with our framework, auto-refocus effect can be {inherently} generated by predicting depth for each ray.}
    \label{fig:teaser}
\end{center}%
}]
 

\begin{abstract}
  In this paper, we present an efficient and robust deep learning solution for novel view synthesis of complex scenes. In our approach, a 3D scene is represented as a light field, i.e., a set of rays, each of which has a corresponding color when reaching the image plane. For efficient novel view rendering, we adopt a two-plane parameterization of the light field, where each ray is characterized by a 4D parameter. We then formulate the light field as a 4D function that maps 4D coordinates to corresponding color values. We train a deep fully connected network to optimize this implicit function and memorize the 3D scene. Then, the scene-specific model is used to synthesize novel views. Different from previous light field approaches which require dense view sampling to reliably render novel views, our method can render novel views by sampling rays and querying the color for each ray from the network directly, thus enabling high-quality light field rendering with a sparser set of training images. Per-ray depth can be optionally predicted by the network, thus enabling applications such as auto refocus. Our novel view synthesis results are comparable to the state-of-the-arts, and even superior in some challenging scenes with refraction and reflection. We achieve this while maintaining an interactive frame rate and a small memory footprint.
\end{abstract}

\section{Introduction}
\label{sec:intro}
Novel view synthesis has long been studied by the computer vision and computer graphics community. It has many applications in multimedia, AR/VR, gaming, etc. Traditional computer vision approaches such as multi-view stereo (MVS) and structure-from-motion (SfM) aim to build a geometric representation of the scene first. An alternative approach is image-based rendering \cite{levoy1996light, lumigraph, unstructuredlumigraph}, where no underlying geometric model or only a simple proxy is needed. These methods can achieve photorealistic rendering. However, a typical light field setup prefers a dense sampling of views around a scene. It thus limits practical use of such an approach.

With the recent advancement of neural rendering \cite{NeuralSTAR2020}, photorealistic rendering with only a sparse set of inputs can be achieved. One approach is to use an explicit geometric representation of a scene reconstructed using a traditional computer vision pipeline and learning-based rendering. Object-specific or category-specific meshes or multi-plane images (MPI) \cite{MPI2018} can be used as the representation. However, these explicit representations do not allow a network to learn the optimal representation of the scene. To achieve this, volume-based representations can be used \cite{Neural_Volumes19}. 
But they typically require a large amount of memory space, especially for complex scenes. 

Memory-efficient implicit representations have gained interests from the research community. For example, surface-based implicit representations can achieve state-of-the-art results and can provide a high-quality reconstruction of the scene geometry \cite{kellnhofer2021neural}. However, surface-based representations face challenges when dealing with complex lighting and geometry, such as transparency, translucency, and thin geometric structures. More recently, volume-based implicit representation achieves remarkable rendering results (e.g., NeRF \cite{mildenhall2020nerf}) and inspires follow-up research. One drawback of NeRF, nevertheless, is the time complexity of rendering. {NeRF, in its original form, needs to query the network multiple times per ray 
and accumulate color and density along the query ray}, which prohibits real-time applications. Although there have been many efforts to accelerate NeRF, they typically require depth proxy to train or rely on additional storage to achieve faster rendering~\cite{hedman2021baking,yu2021plenoctrees,reiser2021kilonerf}.

We propose an efficient novel view synthesis framework, which we call Neural 4D Light Field (NeuLF). We define a scene as an implicit function that maps 4D light field rays to corresponding color values directly, 
This function can be implemented as a Multilayer Perceptron (MLP) and can be learned using only a sparse set of calibrated images placed in front of the scene. This formulation allows the color of a camera ray to be learned directly by the network and does not require a time-consuming ray-marcher during rendering as in NeRF. Thus, NeuLF achieves 1000x speedup over NeRF during inference, while producing similar or even better rendering quality. Our light field setup limits the novel viewpoints to be on the same side of the cameras, e.g., front views only. Despite these constraints, we argue that for many applications such as teleconferencing, these are reasonable trade-offs to gain much faster inference speed with high-quality rendering and a small memory footprint. 

\textbf{Comparison with NeRF}: 
Although our work is inspired by NeRF \cite{mildenhall2020nerf}, there are some key distinctions. NeRF represents the continuous scene function as a 5D radiance field. Such a representation has redundancy, i.e., color along a ray is constant in free space. By restricting the novel viewpoints to be outside of the convex hull of the object, the 5D radiance field can be reduced to a light field in a lower dimension, e.g. 4D. 
Table~\ref{tab:vsnerf}, we summarize the differences between NeuLF and NeRF.


\begin{table}[!h]
\centering
\caption{The comparisons between our proposed NeuLF and NeRF~\cite{mildenhall2020nerf}.\label{tab:vsnerf}}
\resizebox{0.95\columnwidth}{!}{
\begin{tabular}{l|cc}
\hline
                         & NeRF & NeuLF  \\ \hline
Input & $5D$ & $4D$  \\ 
Output & radiance, density & color \\
Viewpoint range   & $360^\circ$ & front views \\ 
Rendering method         & raymarching & direct evalution\\
Rendering speed          & slow & fast \\ 
Memory consumption       & small  & small   \\ 
High-quality rendering   & yes & yes \\\hline
\end{tabular}
}
\end{table}

{Moreover, NeuLF can also optionally estimate per ray depth by enforcing multi-view and depth consistency. Using depth, applications such as auto refocus can be enabled.}  We show state-of-the-art novel view synthesis results on benchmark datasets and our own captured data (Fig.~\ref{fig:teaser}). The comparisons with existing approaches also validate the efficiency and effectiveness of our proposed method. In summary, our contributions are:
\begin{itemize}
    \item {We proposed a fast and memory-efficient novel view synthesis pipeline, which solves the mapping from 4D rays to colors directly.}
    \item {Compared with the state-of-the-arts, our method is better than NeRF and NeX when the scene contains challenging refraction and reflection effects. In addition, our method only needs 25\% of the original input on those challenge scenes to achieve a similar or even better quality.}
    \item {Application wise, the framework we proposed can optionally estimate depth per ray; thus enabling applications such as 3D reconstruction and auto refocus.}
\end{itemize}

\section{Related Work}

Our work builds upon previous work in traditional image-based rendering and implicit-function-based neural scene representation. In the following sections, we will review these fields and beyond in detail.

\textbf{Image-based Rendering}
For novel view synthesis, image-based rendering has been studied as an alternative to geometric methods. In the seminal work of light field rendering \cite{levoy1996light}, a 5D radiance field is reduced to a 4D light field considering the radiance along a ray remains constant in free space. The ray set in a light field can be parameterized in different ways, among which two-plane parameterization is the most common one. Rendering novel views from the light field involves extracting corresponding 2D slices from the 4D light field. To achieve better view interpolation, approximate geometry can be used \cite{lumigraph,unstructuredlumigraph,wood2000surface}. Visual effects of variable focus and variable depth-of-field can also be achieved using light field \cite{ReparameterizedLF}.

With the advancement of deep learning, a few learning-based methods have been proposed to improve the traditional light field. For example, LFGAN \cite{chen2020lfgan} can learn texture and geometry information from light field data sets and in turn predict a small light field from one RGB image. \cite{meng2020high} enables high-quality reconstruction of a light field by learning the geometric features hierarchically using a residual network. 
\cite{wu2021revisiting} integrates an anti-aliasing module in a network to reduce the artifacts in the reconstructed light field. Our method learns an implicit function of the light field and achieves high-quality reconstruction with a sparse input.

\begin{figure*}[!htb]
    \centering
    \includegraphics[width=0.8\textwidth]{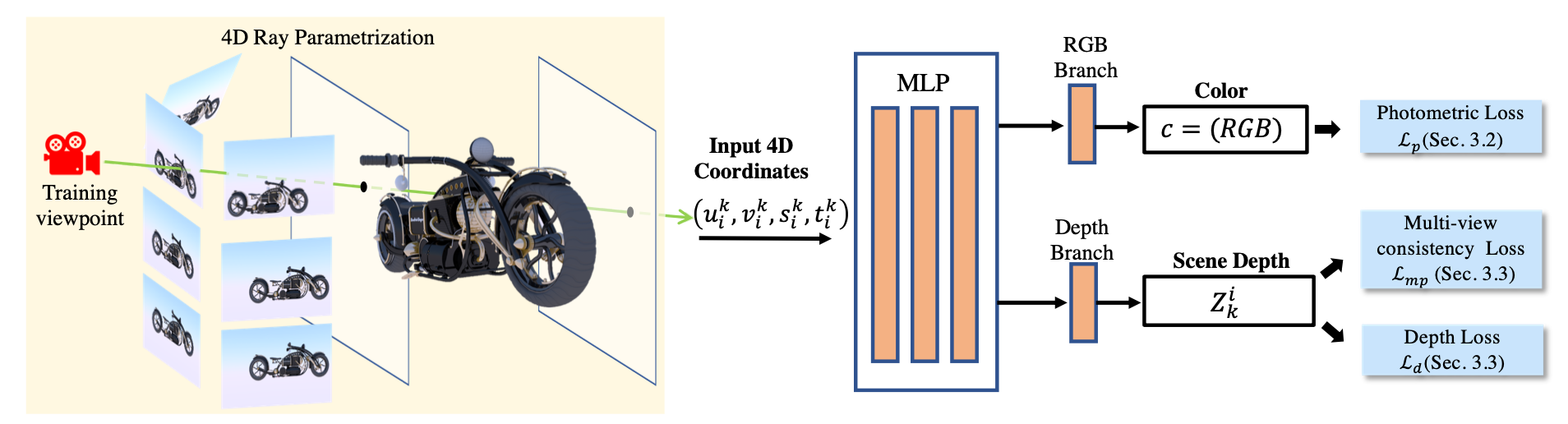}
    \caption{An overview of the Neural 4D Light Field (NeLF). For a set of sampled rays from training images, their 4D coordinates and the corresponding color values can be obtained. The input for NeLF is the 4D coordinate of a ray (query) and the output is its RGB color and scene depth. By optimizing the differences between the predicted colors and ground-truth colors, NeLF can faithfully learn the mapping between a 4D coordinate that characterizes the ray and its color. We also build a depth branch to let the network learn the per ray scene depth by self-supervised losses $\mathcal{L}_{mp}$ and $\mathcal{L}_d$.}
    \label{fig:pipeline}
\end{figure*}

\textbf{Neural Scene Representation}
Neural rendering is an emerging field. One of the most important applications of neural rendering is novel view synthesis. A comprehensive survey of the topic can be found in \cite{NeuralSTAR2020}.

An explicit geometric model can be used as the representation of a scene. \cite{FVS} creates a proxy geometry of the scene using structure from motion and multi-view stereo (MVS). Then, a recurrent encoder-decoder network is used to synthesize new views from nearby views. To improve blending on imperfect meshes from MVS, \cite{DeepBlending} uses predicted weights from a network to perform blending. A high-quality parameterized mesh of the human body \cite{NLT2021} and category-specific mesh reconstruction \cite{kanazawa2018learning} can also be used as the proxy. Recently, Multi-plane Image (MPI) \cite{MPI2018} has gained popularity. \cite{MPI2018} learns to predict MPIs from stereo images. The range of novel views is later improved by \cite{srinivasan2019pushing}. \cite{deepview2019} uses learned gradient descent to generate an MPI from a set of sparse inputs. \cite{mildenhall2019local} uses an MPI representation for turning each sampled view into a local light field. NeX \cite{wizadwongsa2021nex} represents each pixel of an MPI with a linear combination of basis functions and achieves state-of-the-art rendering results in real-time. MPI representation might typically lead to stack-of-cards artifacts. \cite{deepvoxels19} trains a network to reconstruct both the geometry and appearance of a scene on a 3D grid. For dynamic scenes, Neural Volumes (NV) \cite{Neural_Volumes19} uses an encoder-decoder network to convert input images into a 3D volume representation. \cite{lombardi2021mixture} extends NV using a mixture of volumetric primitives to achieve better and faster rendering. While volume-based representations allow for learning the 3D structure, they require large memory space, especially for large scenes. 

Implicit-function-based approaches provide memory-efficient alternatives to explicit representations, while still allowing learning the 3D structure of the scene. Implicit representations can be categorized as implicit surface-based and implicit volume-based approaches.  
SRN~\cite{srn2019} maps 3D coordinates to a local feature embedding at these coordinates. Then, a trained ray-marcher and a pixel generator are used to render novel views. 
IDR~\cite{multiviewneuralsurfacerecon2020} uses an implicit Signed Distance Function (SDF) to model an object on 3D surface reconstruction. Neural Lumigraph \cite{kellnhofer2021neural} provides even better rendering quality by utilizing a sinusoidal representation network (SIREN) to model the SDF. 

Our work is inspired by NeRF \cite{mildenhall2020nerf}, which uses a network to map continuous 5D coordinates (location and view direction) to volume density and view-dependent radiance. Recent works have extended NeRF to support novel illumination conditions \cite{srinivasan2020nerv}, rendering from unstructured image collections from the internet \cite{MartinBrualla20arxiv_nerfw}, large-scale unbounded scenes \cite{zhang2020nerf++}, unknown camera parameters \cite{wang2021nerf}, anti-aliasing \cite{barron2021mip}, deformable models \cite{Park20arxiv_nerfies}, dynamic scenes \cite{li2021neural}, etc. A lot of effort has been put into speeding up rendering with NeRF. DONeRF \cite{neff2021donerf} places samples around scene surfaces by predicting sample locations along each ray. However, transparent objects will pose issues and it requires ground-truth depth for training. FastNeRF \cite{garbin2021fastnerf} achieves 200fps by factoring NeRF into a position-dependent network and a view-dependent network. This allows efficient caching of network outputs during rendering. \cite{yu2021plenoctrees} trains a NeRF-SH network, which maps coordinates to spherical harmonic coefficients and pre-samples the NeRF-SH into a sparse voxel-based octree structure. These pre-sampling approaches sacrifice additional memory storage for speedups. 
NSVF \cite{Liu20neurips_sparse_nerf} represents a scene using a set of NeRF-like implicit fields defined on voxels and uses a sparse octree to achieve 10x speedup over NeRF during rendering. KiloNeRF  \cite{reiser2021kilonerf} decomposes a scene into a grid of voxels and uses a smaller NeRF for each voxel. Storage costs will increase when more networks are used. Using AutoInt \cite{autoint2021}, calculations of any definite integral can be done in two network evaluations; this achieves 10x acceleration, but rendering quality is decreased. Compared with these approaches, our method achieves 1000x speedup over NeRF by representing the scene with an implicit 4D light field without any additional pre-sampling or storage overhead.

{Recently a concurrent work \cite{sitzmann2021lfns} propose to use a network to direct regress the mapping from the 6D Plücker coordinates to colors. It leverages meta-learning to enables view synthesis using a single image observation in Shape{N}et dataset~\cite{chang2015shapenet}. In contrast, we use 4D representation and conduct extensive experiments on real-world scenes. Another concurrent work \cite{feng2021signet} transforms 4D light field representation by leveraging Gegenbauer polynomials basis, and learning the mapping from this basis function to color, however, it requires dense narrow baseline input with planar camera arrangement. 


\section{Our Method}

In Fig.~\ref{fig:pipeline}, we illustrate the pipeline of our system. In the following sections, we will first briefly discuss the light field, followed by our NeuLF representation and the proposed loss functions. We will also discuss our training strategies. 

\subsection{4D Light Field Representation}

All possible light rays in a space can be described by a 5D plenoptic function. Since radiance along a ray is constant if viewed from outside of the convex hull of the scene, this 5D function can be reduced to a 4D light field \cite{levoy1996light,lumigraph}. The most common parameterization is a two-plane model shown in Fig.~\ref{fig:4dlf}. Each ray from the camera to the scene will intersect with the two planes. Thus, we can represent each ray using the coordinates of the intersections, $(u,v)$ and $(s,t)$, or simply a 4D coordinate $(u,v,s,t)$. Using this representation, all rays from the object to one side of the two planes can be uniquely determined by a 4D coordinate. 

\begin{figure}[!b]
    \centering
    \includegraphics[width=0.85\linewidth]{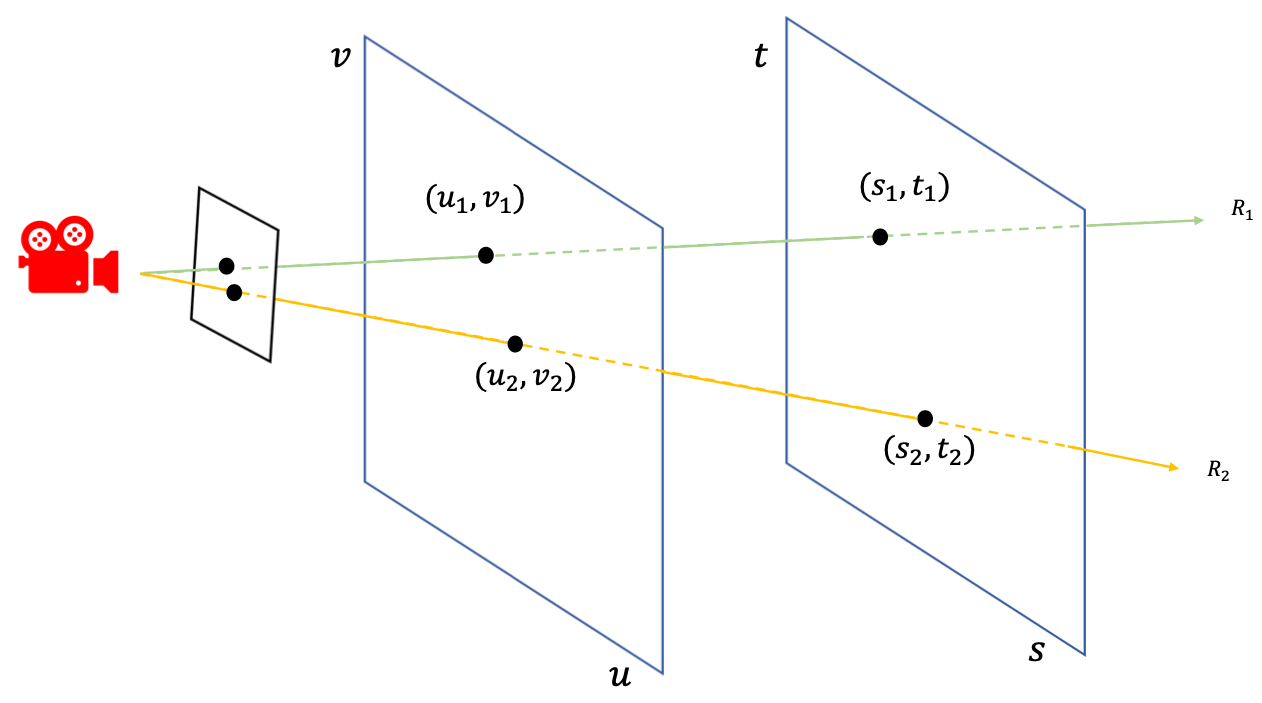}
    \caption{The 4D light field representation. Each ray is characterized by 4 parameters $(u,v)$ and $(s,t)$, which uniquely locates the ray.}
    \label{fig:4dlf}
\end{figure}

Based on this representation, rendering a novel view can be done by querying all the rays from the center of projection to every pixel on the camera's image plane. We denote them as $\{R_1,R_2,...,R_N\}$, where $N$ is the total number of pixels. Then, for the $i$-th ray $R_i$, we can obtain its 4D coordinate $(u_i,v_i,s_i,t_i)$ by computing its intersections with the two planes. If a function $f$ maps the continuous 4D coordinates to color values, we can obtain the color of $R_i$ by evaluating the function $f_c(u_i,v_i,s_i,t_i)$. In the next section, we will introduce Neural 4D Light Field (NeuLF) for reconstructing this mapping function $f$. 

\subsection{Neural 4D Light Field Reconstruction}

We formulate the mapping function $f_c$ as a Multilayer Perceptron (MLP). The input of this MLP is a 4D coordinate and the output is RGB color. As shown in Fig.~\ref{fig:pipeline}. The goal of the network is to learn the mapping function from training data.

\textbf{Training Data}: for a given scene, the training data comes from a set of captured images $\{I_1,I_2,...,I_M\}$, where $M$ is the total number of images. Assuming the camera pose for each image is known or obtainable, for each image $I_k (k=1,...,M)$, we can traverse its pixels and generate all corresponding rays $\{R_1^k,R_2^k,...,R_{N_k}^k\}$, where $N_k$ is the total number of pixels in the $k$-th image. Based on the 4D 
light field representation, all 4D coordinates
$\left\{\left(u_1^k,v_1^k,s_1^k,t_1^k\right),...,\left(u_{N_k}^k,v_{N_k}^k,s_{N_k}^k,t_{N_k}^k\right)\right\}, (k=1,...,M)$, can be obtained. On the other hand, the color for each pixel is known from the input images. To this end, we have constructed a collection of sample mappings from 4D coordinates to color values $\left(u_i^k,v_i^k,s_i^k,t_i^k\right) \rightarrow c_i^k, (k=1...M, i=1...N_k)$, where $c_i^k$ is the color of the $i$-th pixel on the $k$-th image. By feeding this training data to the MLP network, the parameters $\Theta$ can be learned by minimizing the following photometric loss $\mathcal{L}_p$:


\begin{equation}
    \label{eqn:photometricloss}
    \mathcal{L}_p = \sum_{k=1}^{M}\sum_{i=1}^{N_k}\norm{f_c\left(u_i^k,v_i^k,s_i^k,t_i^k \given \Theta\right)-c_i^k}_2
\end{equation}

In Fig.~\ref{fig:pipeline}, we demonstrate an example of capturing images to train our neural 4D light field representation with a set of  unstructured front-faced camera views. In this example, the cameras are placed on one side of two light slabs. 
 


\textbf{Rendering}: Given a viewpoint $\mathcal{V}$, we can render a novel view $\mathcal{R}\left(\mathcal{V}\right)$ by evaluating the learned mapping function $f$. With the camera pose and the desired rendering resolution $\{W^\mathcal{V},H^\mathcal{V}\}$, we sample all rays $\{R_1^\mathcal{V},R_2^\mathcal{V},...,R_{N_\mathcal{V}}^\mathcal{V}\}$, where $N_\mathcal{V}=W^\mathcal{V}\times H^\mathcal{V}$ is the number of pixels to be rendered. We can further calculate the 4D coordinates $\left\{\left(u_i^\mathcal{V},v_i^\mathcal{V},s_i^\mathcal{V},t_i^\mathcal{V}\right)\right\}$ for each ray $R_i^\mathcal{V}, (i=1...N_\mathcal{V})$.
We then formulate the rendering process $\mathcal{R}$ as $N_\mathcal{V}$ evaluations of the mapping function $f_c$:

\begin{equation}
    \label{eqn:rendering}
    \mathcal{R}\left(\mathcal{V}\right) = \left\{f_c\left(u_i^\mathcal{V},v_i^\mathcal{V},s_i^\mathcal{V},t_i^\mathcal{V} \given \Theta\right), i=1,...,N_\mathcal{V}\right\}.
\end{equation}

\begin{figure}
    \centering
    \includegraphics[width=0.8\linewidth]{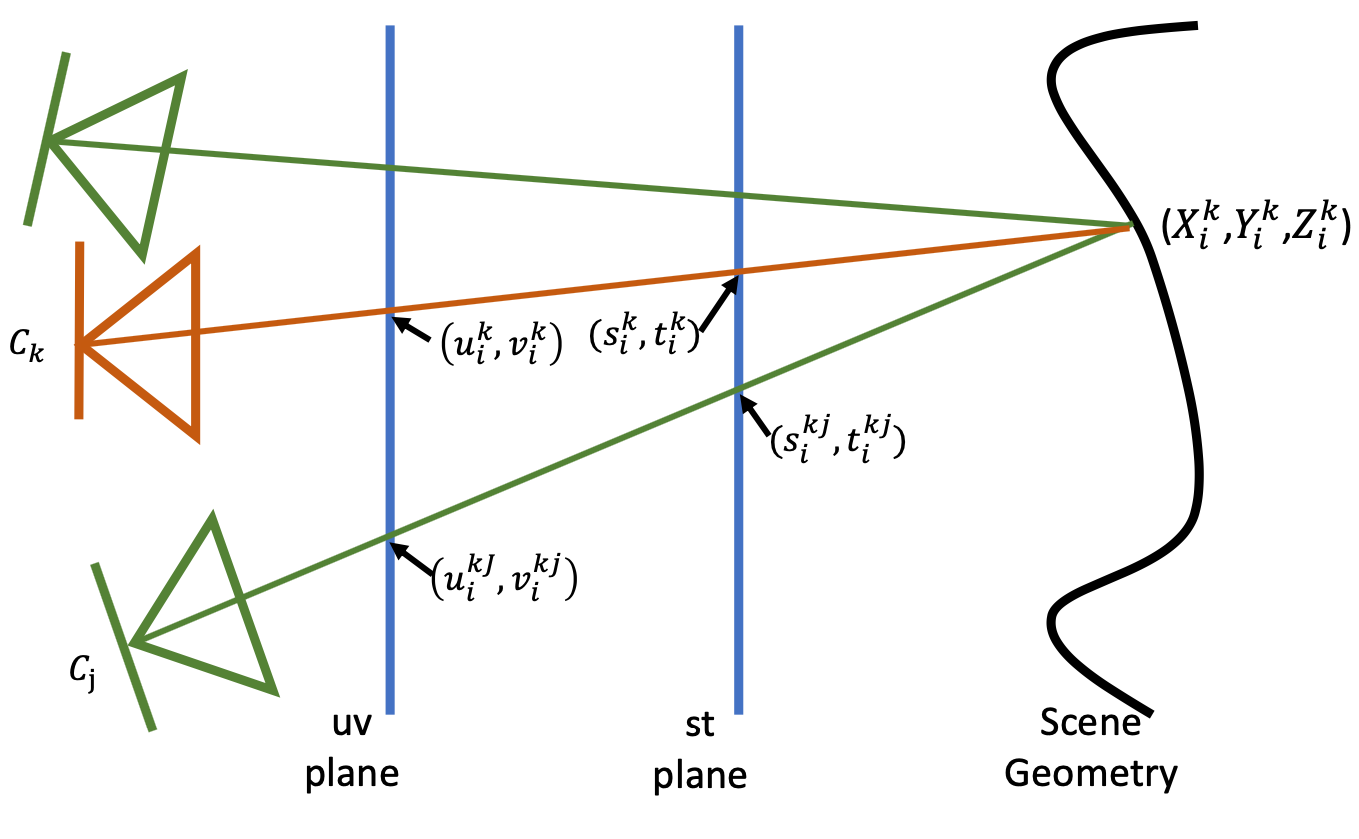}
    \caption{Given the 4D query ray $(u_i^k,v_i^k,s_i^k,t_i^k)$ (red line) and its predicted depth $Z_i^k$, we compute its corresponding rays (green line) to its nearest camera views $C_i$, which are used for self-supervision.}
    \label{fig:Depth_fig}
\end{figure}


\subsection{4D Light Field Scene Depth Estimation}
To simulate variable focus and variable depth-of-field using light field, previous work~\cite{ReparameterizedLF} dynamically reparameterizes the light field by manually moving the focal surface as a plane. To optimally and automatically select a focal plane given a pixel location, we aim to solve the focal surface manifold geometry that conforms to the scene geometry \zhongli{under Lambertian scene assumption}. 

\xuyi{We let the same MLP predict per-ray depth as shown in Fig.~\ref{fig:pipeline}.} As in Fig.~\ref{fig:Depth_fig}, for a 4D query ray $(u_i^k,v_i^k,s_i^k,t_i^k)$ from the camera $C_k$, the predicted scene depth is $Z_i^k = f_d(u_i^k,v_i^k,s_i^k,t_i^k\given\Theta)$ where $f_d$ is the network mapping function from 4D coordinates to scene depth. We compute \xuyi{ray-surface} intersection $(X_i^k,Y_i^k,Z_i^k)$. Then, we self-supervise $Z_i^k$ by applying multi-view consistency cues. To do this, we trace the rays from $(X_i^k,Y_i^k,Z_i^k)$ back to $C_k$'s \textit{K}-nearest data cameras $\{C_j\},(j=1,...,K)$ (\textit{K}=5 in our experiments). Those rays intersect \textit{uv}-\textit{st} plane and can be parameterized as: 

\begin{equation}
        f_{uvst}(i,j,k,Z_i^k) = \{u_{i}^{kj},v_{i}^{kj},s_{i}^{kj},t_{i}^{kj}\}
\end{equation}

Ray-plane intersection is differentiable. We propose two loss functions $\mathcal{L}_{mp}$ and $\mathcal{L}_d$ to minimize the multi-view photometric error and depth differences respectively as follows:

\begin{equation}
    \label{eqn:loss_mp}
    \mathcal{L}_{mp} = \norm{f_c\left(u_i^k,v_i^k,s_i^k,t_i^k \given \Theta\right)- \mathcal{C}_{i}^k(Z^k_i)}_2
\end{equation}

\begin{equation}
    \label{eqn:super-color}
    \mathcal{C}_{i}^k(Z^k_i) = \sum_{j=1}^K\sum_{i=1}^{N_k}\omega(i,j)f_c\left(f_{uvst}(i,j,k,Z_i^k) \given \Theta\right)
\end{equation}

\begin{equation}
    \label{eqn:loss_d}
    \mathcal{L}_{d} = \norm{f_d\left(u_i^k,v_i^k,s_i^k,t_i^k \given \Theta\right)- \mathcal{D}_{i}^k(Z^k_i)}_2
\end{equation}

\begin{equation}
    \label{eqn:super-depth}
    \mathcal{D}_{i}^k(Z^k_i) = \sum_{j=1}^K\sum_{i=1}^{N_k}\omega(i,j)f_d\left(f_{uvst}(i,j,k,Z_i^k) \given \Theta\right)
\end{equation}

where $\omega(i,j)$ are the normalized weights $\omega(i,j) = \frac{\frac{1}{d^2_{ij}}}{\sum_j \frac{1}{d^2_{ij}}}$ with the Euclidean distance  between the camera $i$ and its neighbor camera $j$ as $d_{ij}$. $\mathcal{C}_{i}^k(Z^k_i)$ and $\mathcal{D}_{i}^k(Z^k_i)$ are the supervise color and depth sum over weighted nearest data cameras with network output $Z^k_i$ respectively. Training with both $\mathcal{L}_{mp}$ and $\mathcal{L}_{d}$ will encourage the MLP to learn the depth representation of the scene.


\begin{figure}[!ht]
    \centering
    \includegraphics[width=0.75\linewidth]{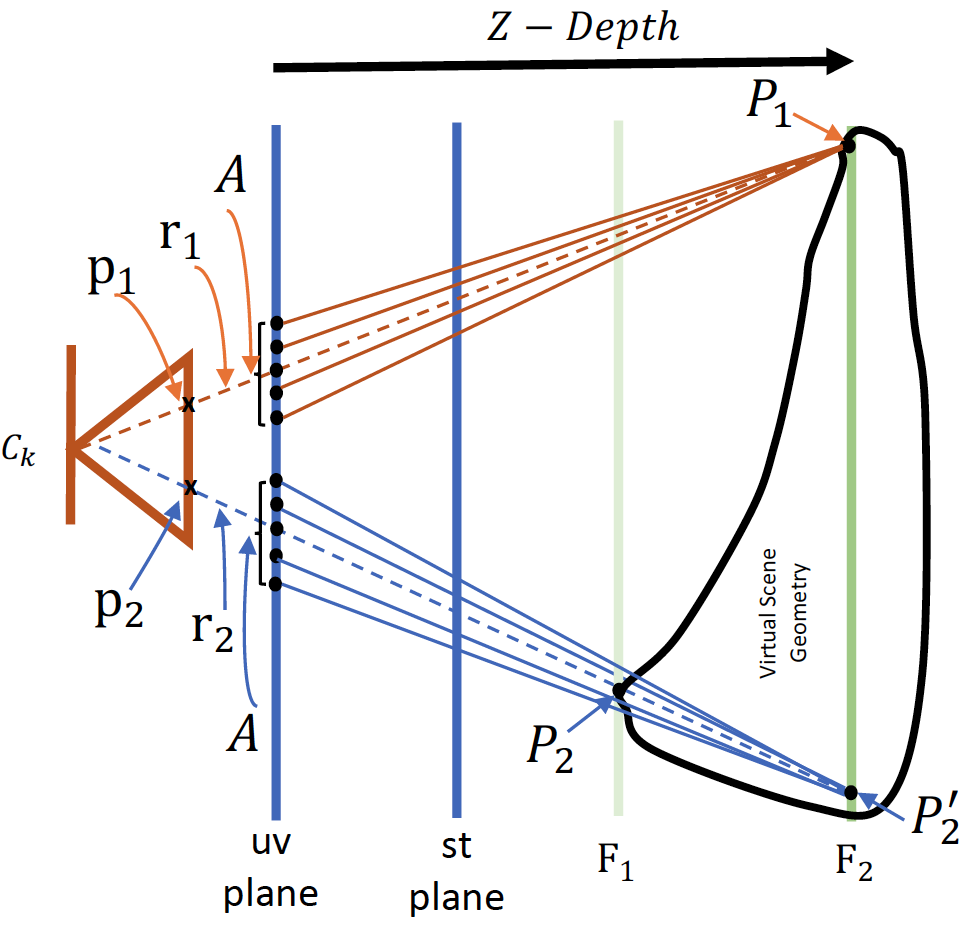}
    \caption{Auto-refocus illustration. Reconstruct two rays $r_1$ and $r_2$ by combining ray's within their aperture A. When focal plane correspond to $r_1$'s depth $P_1$, the $r_1$ will appears in focus, while $r_2$ will not.}
    \label{fig:refocusTech}
\end{figure}

\xuyi{With ray-based depth, we can enable efficient auto-refocus effect by adopting a dynamic parameterization of the light field ~\cite{ReparameterizedLF}. More specifically, we simulate the depth-of-field effect by combining rays within an given aperture size. Fig.~\ref{fig:refocusTech} shows the case of two rays $r_1$ and $r_2$. The two rays intersect camera image plane at pixels $p_1$ and $p_2$, intersect the scene geometry at $P_1$ and $P_2$, and intersect a given focal plane $F_2$ at $P_1$ and $P'_2$. To reconstruct the final color of the pixels $p_1$ and $p_2$, we collect a cone of sample rays originated from $P_1$ and $P'_2$ on the focal plane $F_2$ within an aperture $A$. We then query the network $f_c$ to obtain the ray colors. These ray colors are weighted-averaged to produce the final pixel color. In this case, $P_1$ is on the surface of the object, while $P'_2$ is not. Thus, the image pixel $p_1$ appears in focus, while pixel $p_2$ is blurred since it combines colors from a small area around surface point $P_2$. }

\xuyi{To auto-refocus at pixel location $p_2$, we extract its depth by query the depth network, and set a new focal plane $F_1$ at pixel $p_2$'s depth. Rendering the {N}eu{LF} with this new focal plane will make pixel $p_2$ in focus while blur pixel $p1$.The results are shown in Fig.~\ref{fig:refocus}}.



\begin{figure}[!ht]
    \centering
    \includegraphics[width=\linewidth]{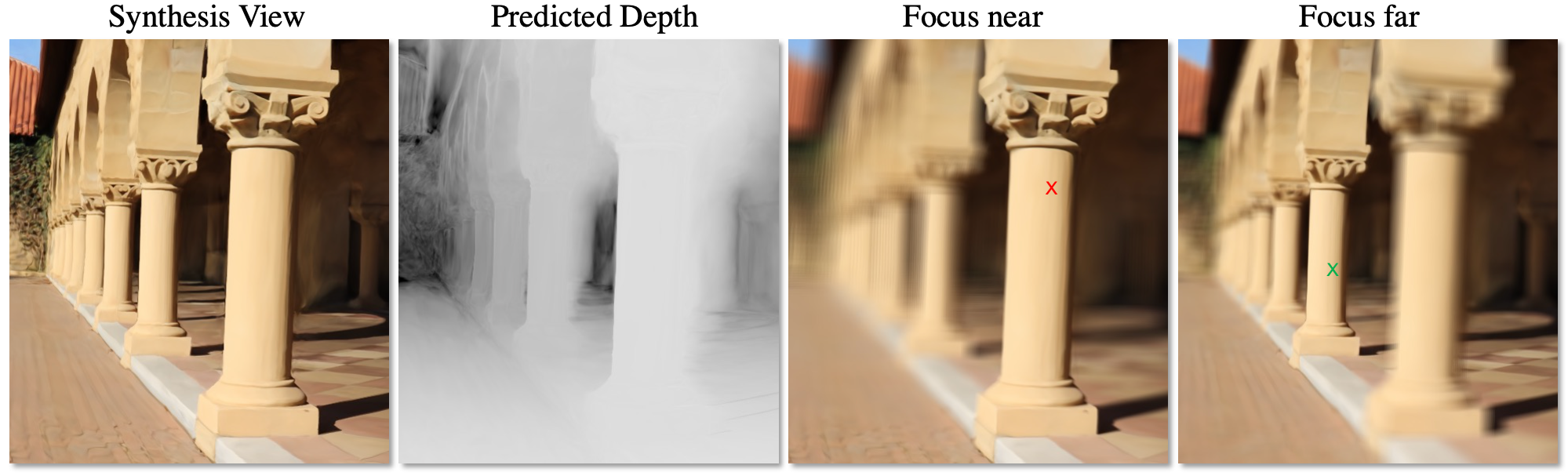}
    \caption{Depth estimation result and refocus effects. We show a novel view and its corresponding depth estimation.  In addition, we show two auto-refocus results with focus point on the red cross (near) and  on the green cross (far) given by a user click.}
    \label{fig:refocus}
\end{figure}

\section{Experimental Results}

We first discuss the implementation details of NeuLF. Then we perform quantitative and qualitative evaluations against state-of-the-art methods for novel view synthesis. 

\begin{figure*}
    \centering
    \includegraphics[width=0.85\linewidth]{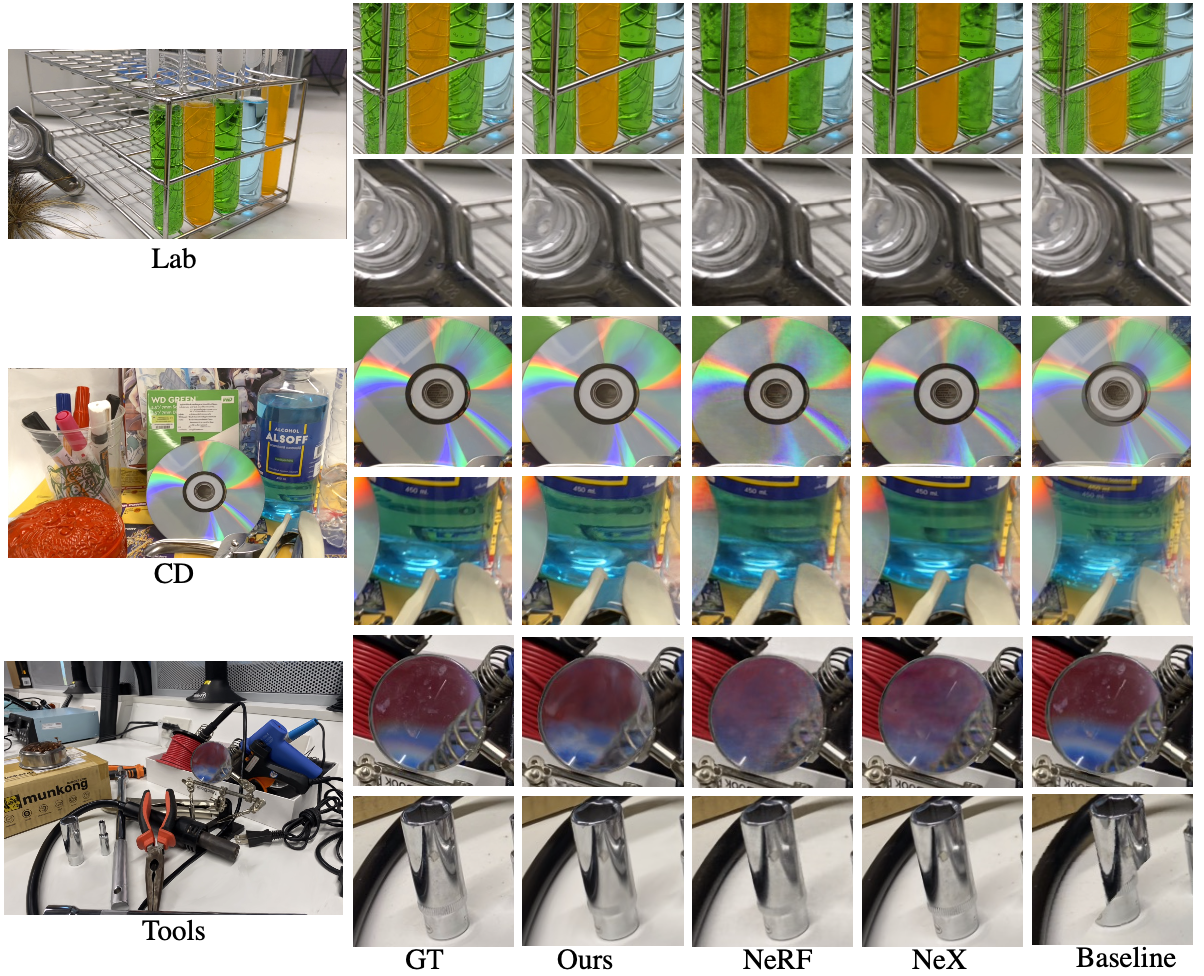}
    \caption{Qualitative Results on test views from shinny dataset. Our method captures more details on the reflection and refraction areas of the scenes. }
    \label{fig:quantitative}
\end{figure*}

\addtolength{\tabcolsep}{-2.5pt}    
\begin{table*}[!h]
\centering
\caption{Average scores across test views for each scene in the shinny dataset.\label{tab:shinny}}
\resizebox{0.85\textwidth}{!}{%
\begin{tabular}{l|cccc|cccc|cccc}
\hline
& \multicolumn{4}{c|}{PSNR$\uparrow$} & \multicolumn{4}{c|}{SSIM$\uparrow$} & \multicolumn{4}{c}{LPIPS$\downarrow$} \\
& Baseline & NeX  & NeRF  & Ours  & Baseline & NeX  & NeRF  & Ours  & Baseline & NeX  & NeRF  & Ours\\ \hline
Lab            & 21.69 & 30.43 & 29.60 & \textbf{31.95}  & 0.693 & 0.949 & 0.936 & \textbf{0.951} & 0.261 & 0.146 & 0.182 &\textbf{0.097}   \\
CD            & 20.70 & 31.43 & 30.14 & \textbf{32.11} & 0.551 & 0.958 & 0.937 & \textbf{0.964} & 0.294 & 0.129 & 0.206 & \textbf{0.123}   \\
Giants     & 16.20 & \textbf{26.00} & 24.86 & 24.95 & 0.265 & \textbf{0.898} & 0.844 & 0.839 & 0.274 & \textbf{0.147} & 0.270 & 0.299      \\
Tools     & 16.19 & \textbf{28.16} & 27.54 & 26.73 & 0.575 & \textbf{0.953} & 0.938 & 0.896 & 0.250 & \textbf{0.151} & 0.204 & 0.167      \\
Food         & 14.57 & \textbf{23.68} & 23.32 & 22.61 & 0.297 & \textbf{0.832} & 0.796 & 0.776 & 0.341 & \textbf{0.203} & 0.308 & 0.322     \\
Pasta     & 12.50 & \textbf{22.07} & 21.23 & 20.64 & 0.216 & \textbf{0.844} & 0.789 & 0.715 & 0.271 & \textbf{0.211} & 0.311 & 0.283      \\
Seasoning       & 17.31 & \textbf{28.60} & 27.79 & 27.12 & 0.412 & \textbf{0.928} & 0.898 & 0.881 & 0.279 & \textbf{0.168} & 0.276 & 0.263      \\
Crest           & 15.91 & \textbf{21.23} & 20.30 & 20.11 & 0.209 & \textbf{0.757} & 0.670 & 0.653 & 0.304 & \textbf{0.162} & 0.315 & 0.410      \\ \hline
\end{tabular}
}
\end{table*}

\subsection{Implementation Details}
\label{sec:training}
We train the MLP on the following overall loss function:

\begin{equation}
    \label{eqn:totalloss}
    \mathcal{L} = \mathcal{L}_p + \lambda_s\mathcal{L}_{mp} + \lambda_r\mathcal{L}_d,
\end{equation}
where the weighting coefficients are $\lambda_s=0.5$ and $\lambda_r=0.1$. These parameters are fine-tuned by mixing the manual tuning and grid search tuning. This set of parameters works best for most of the scenes we tested. \xuyi{For scenes with strong view-dependent effects such as specularities, Lambertian assumption is no longer valid. Therefore, we disable the multi-view photometric error and depth difference terms in the loss for such scenes.} {Grid search can be used to automatically find optimal parameters for a specific scene but will lead to a longer training time.}

{To extract camera rays from input photos, we calibrate the camera poses and intrinsic parameters using a structure-from-motion tool from COLMAP~\cite{schonberger2016structure}.}
During training, we randomly select a batch of camera rays from the training set at each iteration. By passing them to the MLP to predict the color of each ray, we calculate $\mathcal{L}_p$ and back-propagate the error.


{The input 4D coordinate $(u,v,s,t)$ (normalized to $[-1,1]$) is passed through 20 fully-connected ReLU layers, each with 256 channels. We developed a structure that includes a skip connection that concatenates the input 4D coordinate to every 4 layers start with the fifth layer. An additional layer outputs 256-dimensional feature vector. This feature vector is split into the color branch and depth branch. \xuyi{Each branch is followed} with an additional fully-connected ReLU layer with 128 channels, and output 3 channel RGB radiance with sigmoid activation and 1 channel scene depth with sigmoid activation, respectively. For model training, we set the ray batch size in each iteration to $8192$. We train the MLP for 1500 epochs using the Adam optimizer~\cite{kingma2014adam}. The initial learning rate is $5\times 10^{-4}$ and decays 
by $0.995$ every epoch. To train the NeuLF on a scene with 30 input images with a resolution of $567\times 1008$, it takes 5 hours using 1 Nvidia RTX 3090 card. For testing on the same situation, rendering an image costs about 70ms while NeRF takes 51 seconds.} 


\subsection{Comparison with State-of-the-Art Methods}

In this section, we demonstrate the qualitative results of novel view synthesis and compare them with current top-performing approaches: NeRF~\cite{mildenhall2020nerf}, NeX~\cite{wizadwongsa2021nex} and the baseline light field rendering method~\cite{levoy1996light}. We evaluate the models on the shinny dataset~\cite{wizadwongsa2021nex}. In this dataset, each scene is captured by a handheld smartphone in a forward-facing manner with a resolution $567 \times 1008$ or $756 \times 1008$. This is a challenging dataset that contains complex scene geometry and various challenging view-dependent effects. For example, refraction through the test tubes (filled with liquid) and magnifier, rainbow effect emits by a CD disk and sharp specular highlights from silverware and thick glasses.

We hold out $\frac{1}{8}$ of each scene as the test set and the rest of them as the training set. The qualitative results are shown in Figure.~\ref{fig:quantitative}. The leftmost column shows our results on the test view of three challenging scenes (Lab, CD, and Tools). We have zoomed in on the part of the image areas for comparison with other methods. Our method is superior when a scene contains detailed refraction and reflection.  Our result has less noise on glass sparkles. In the CD scene, our result shows more sharp and vivid detail on the rainbow, plastic cup reflections, and less noise on the liquid bottle than NeX and NeRF. In the Tools scene, although our result is not as sharp as the ground truth, it contains more overall details than NeX and NeRF and is able to recover metallic reflection with less noise than others. Our method essentially relies on ray interpolation rather than volumetric rendering like NeRF/NeX. We believe the traditional ray interpolation handles refraction and reflection better than volumetric representation.

The baseline Light Field rendering (last column) exhibits good results when rays are sufficiently sampled (magnifier). However, its method exhibits aliasing and misalignment artifacts in the low ray sampling area (metallic, tube). 

We report three metrics: PSNR (Peak Signal-to-Noise Ratio, higher is better), SSIM (Structural Similarity Index Measure, higher is better), and LPIPS (learned perceptual Image Patch Similarity, lower is better) to evaluate our test results. In Table~\ref{tab:shinny}, we report the three metrics for the 8 scenes in the shinny dataset. We use the NeX and NeRF scores originally report from the NeX paper. For each scene, we calculate the scores by averaging across the views in the test split. Our method produces the highest score across all three metrics on CD and Lab scenes which contain challenging refraction and reflection. For the rest scenes, while NeX has the highest score by producing high-frequency details in the richly-textured area, we generate comparable scores with NeRF. Note that our rendering speed is $1000 \times$ than NeRF.



\subsection{Ablation Study}
\label{sec:ablation1}
\zhongli{To demonstrate the effectiveness of our proposed multiview and depth regularization terms, we use two mostly Lambertian scenes the $columns$ and the $tribe$, which contain complex local details, for ablation studies. We demonstrate how the multi-view and depth loss $\mathcal{L}_{mp}$ and $\mathcal{L}_{depth}$ affect the quality of synthesized views. We show the results of two different scenes. We compared the model trained with and without our proposed regularization terms (MPVL). As shown in Table.~\ref{fig:ablation_fig} and Fig.~\ref{tab:ablation_fig}, our model with multi-view and depth regularization improves the result both qualitatively and quantitatively in both scenes, of which more details can be reconstructed.  For Lambertian scenes, the network trained with multi-view and depth consistency constraints takes the scene geometry into account. Hence the results in unseen views are more reasonable and detail preserved.}

\begin{figure}[!h]
    \centering
    \includegraphics[width=\linewidth]{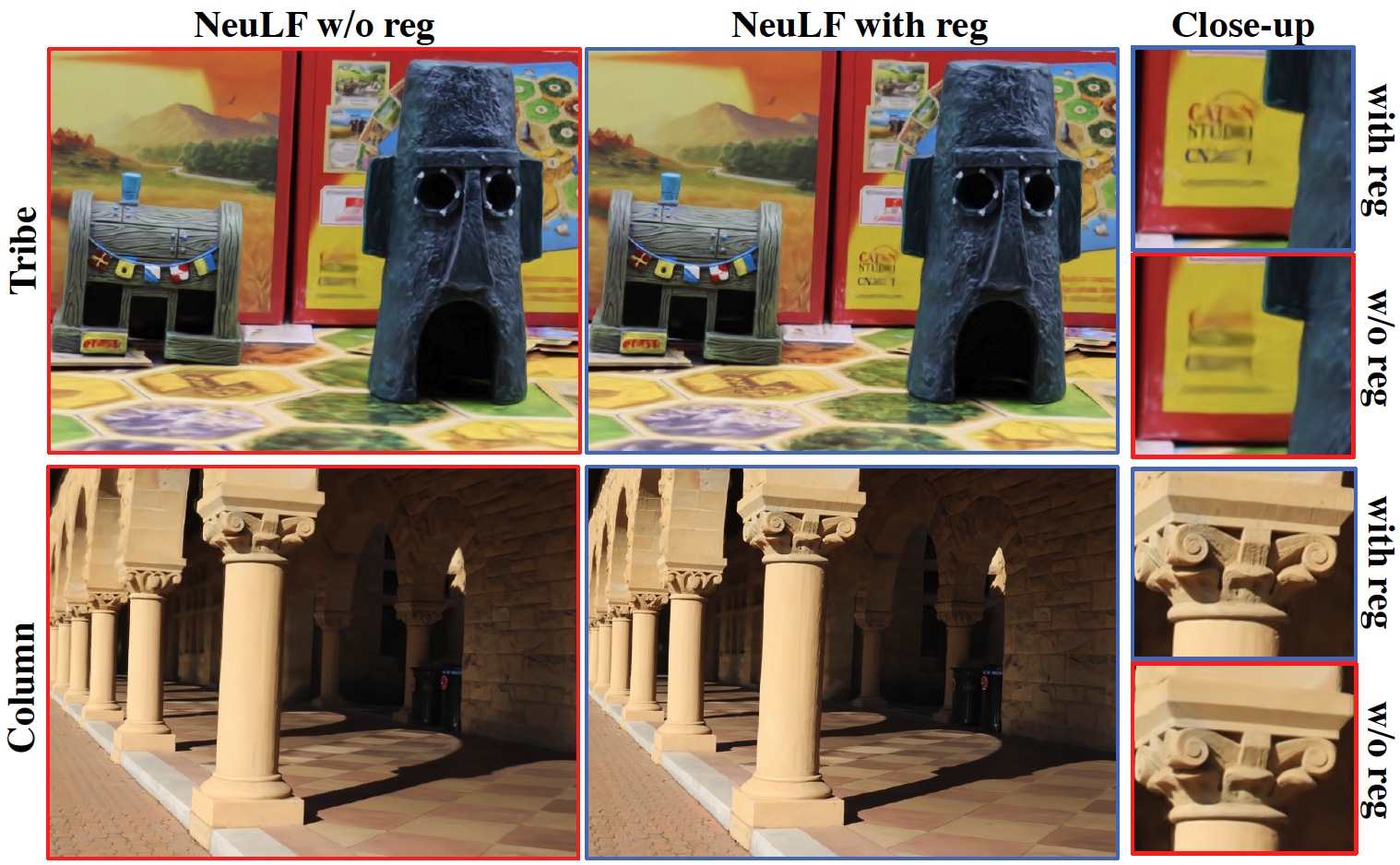}
    \caption{Ablation study. We show results of NeuLF trained with and without multi-view and depth regularization on two Lambertian scenes. The last column shows the close-up comparison of the two scenes.}
    \label{fig:ablation_fig}
\end{figure} 

\begin{table}[!h]
\centering
\caption{A quantitative comparison of NeuLF trained with and without MPDL.
The scores are computed cross test views of Tribe and Column scene.\label{tab:ablation_fig}}
\resizebox{\columnwidth}{!}{%
\begin{tabular}{l|cc|cc|cc}
\hline
& \multicolumn{2}{c|}{PSNR$\uparrow$} & \multicolumn{2}{c|}{SSIM$\uparrow$} & \multicolumn{2}{c}{LPIPS$\downarrow$} \\
& w/o MVDL & w MVDL &  w/o MVDL & w MVDL & w/o MVDL & w/ MVDL\\ \hline
Tribe            & 25.62 & \textbf{26.93} & 0.788 & \textbf{0.830} & 0.232 & \textbf{0.219} \\
column           & 28.58 & \textbf{29.20} & 0.838 & \textbf{0.847} & 0.227 & \textbf{0.211} \\\hline
\end{tabular}
}
\end{table}

\begin{table}[!h]
\centering
\caption{Impacts from the number of inputs. We input different numbers of images to evaluate the performance of our model. \label{tab:ablation}}
\resizebox{\columnwidth}{!}{%
\begin{tabular}{l|ccccc}
\hline
        & method & \#Images &  PSNR $\uparrow$  &  SSIM $\uparrow$ &  LPIPS $\downarrow$  \\ \hline
CD 75\%   & Ours & 230       & 31.81      & 0.959       & 0.126    \\ 
CD 50\%   & Ours      & 153       & 31.41     & 0.953       & 0.145     \\ 
CD 25\%   & Ours      & 77       & 30.16     & 0.948       & 0.170     \\ 
CD 100\% & NeRF       & 307       & 30.14      & 0.937       & 0.206    \\
CD 100\%  & NeX     & 307       & 31.43      & 0.958       & 0.129    \\\hline 
Lab 75\% & Ours      & 227       & 31.87     & 0.949       & 0.097   \\ 
Lab 50\% & Ours      & 151       & 31.74     & 0.948       & 0.104   \\ 
Lab 25\%& Ours       & 76        & 30.61       & 0.939      & 0.116   \\ 
Lab 100\% & NeRF       & 303        & 29.60       & 0.936      & 0.182   \\
Lab 100\% & NeX      & 303        & 30.43       & 0.949      & 0.146   \\\hline
\end{tabular}
}
\end{table}

\subsection{Study on Number of Inputs}
\label{sec:ablation}
To understand how the number of input views affect the novel view synthesis result, we train our model on fewer images. As shown in Figure~\ref{fig:ablation}, we use 75\%, 50\%, and 25\% of the original data for the experiment. Although the input images are dramatically decreased, our method still generates high-quality results. Our results still retain the rainbow and background reflections, and the refraction details through the test tube are also well retained. In Table ~\ref{tab:ablation}, note that even \textit{with less input images for CD and Lab scenes}, our results are still comparable with or even better than NeRF and NeX with full number of inputs in the above challenging scenes. 

\begin{figure}[!h]
    \centering
    \includegraphics[width=\linewidth]{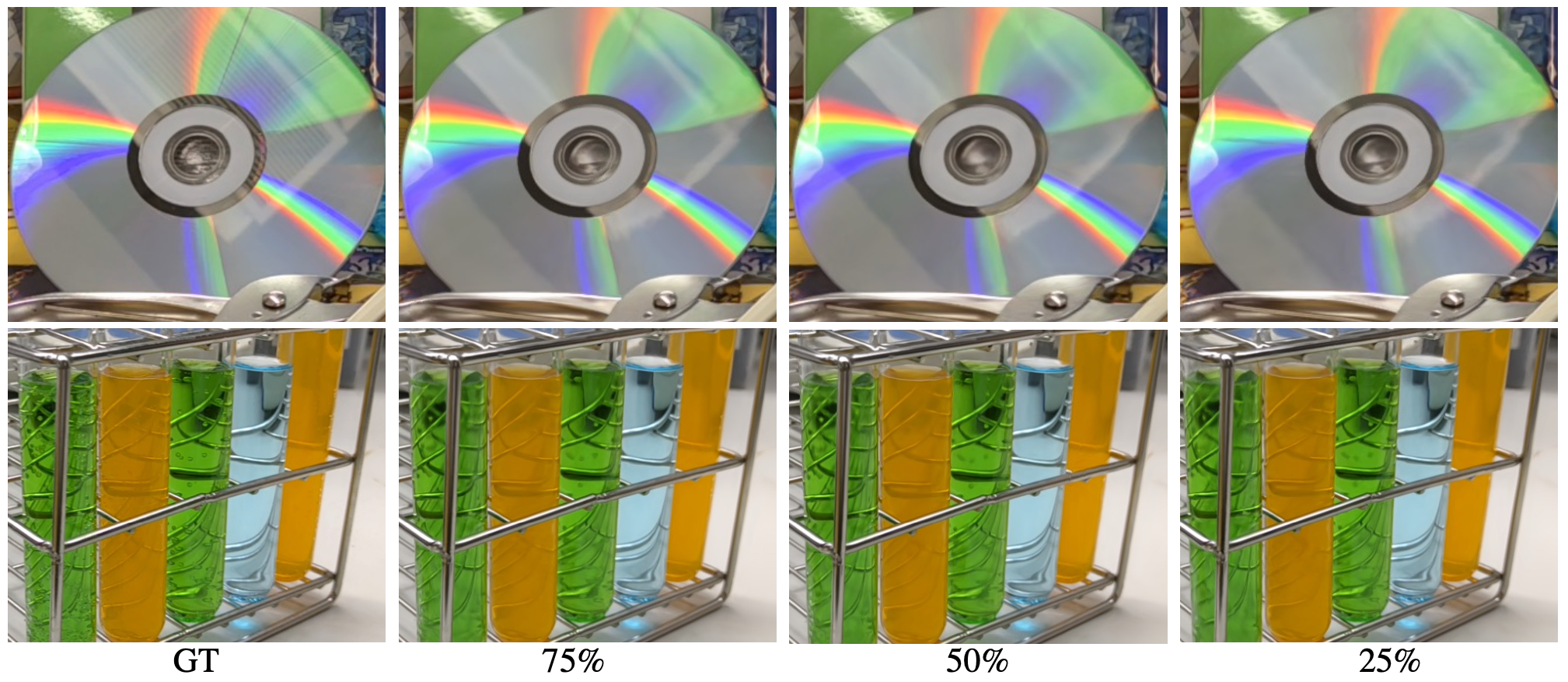}
    \caption{Study on number of inputs. As shown in the figure, we input 75\%, 50\%, and 25\% of the original images on the CD and Lab scenes.}
    \label{fig:ablation}
\end{figure} 


\subsection{Applications}
\label{sec:applications}
As applications of NeuLF, we show results of depth estimation and automatic refocusing.

In Figure~\ref{fig:result_depth}, we show an example of the depth estimation and refocusing effect on our own captured scene Tribe. Note the detailed depth of the house and statue are successfully recovered. With free-viewpoint scene depth, we can automatically select a focal plane given an image pixel location. We show two synthesized novel views rendered from our own captured scene. Then, we show the auto-refocus result given two positions on the image, one focuses on the near object (red cross), and another on the far object (green cross). This is enabled naturally by the dynamic 4D light field representation~\cite{ReparameterizedLF}.
\begin{figure}[!h]
    \centering
    \includegraphics[width=\linewidth]{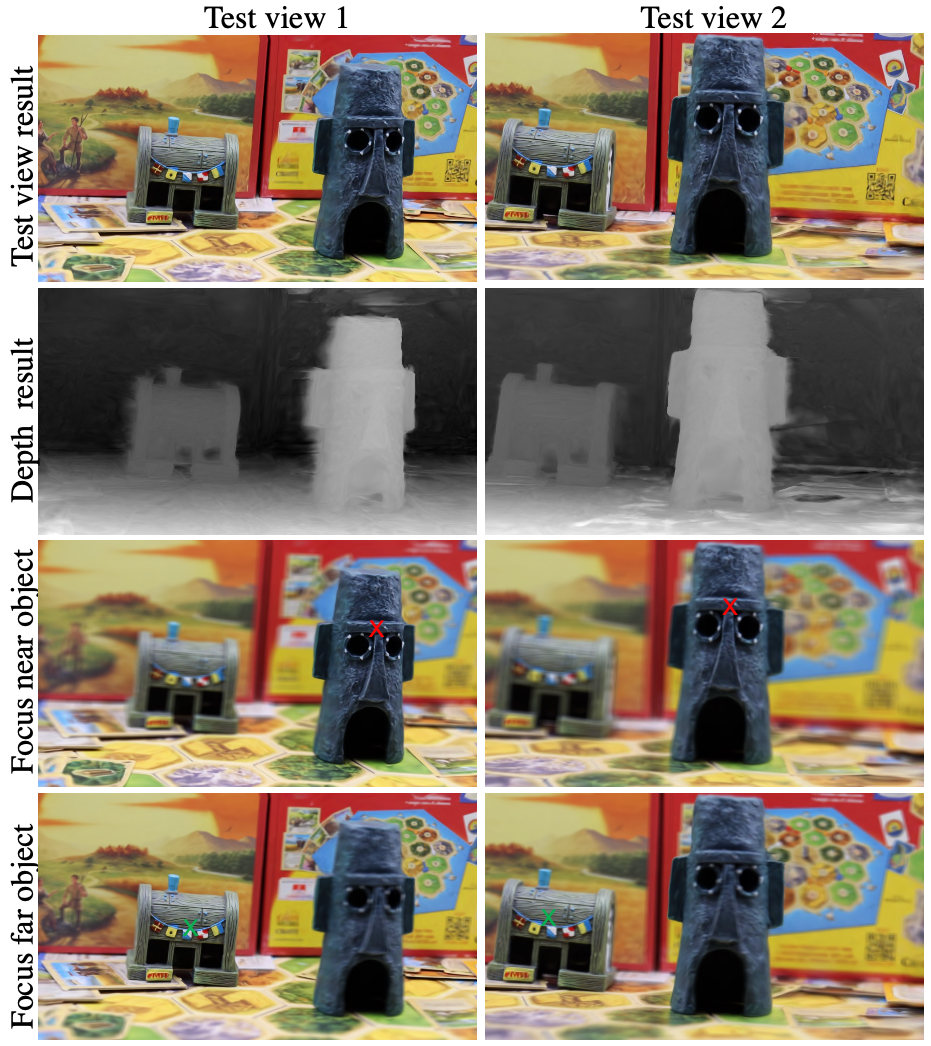}
    \caption{Depth estimation and auto-refocus result. We show the results of two novel views (first row) and the depth estimation (second row). In the third row we refocus on the near object, and in the fourth row we refocus on the far object.}
    \label{fig:result_depth}
\end{figure}

\subsection{Failure Cases}
Our method is based on a 4D 2PP (two-plane parameterization) light field representation. Since each 3D world position corresponds to multiple discontinued 4D coordinates, NeuLF representation is difficult to learn. Therefore, we observe that our model cannot fully recover the high-frequency details in the scene as shown in Figure.~\ref{fig:fail}. The seaweed texture (first row) and object's Hollow-out structure (second row) are over-smoothed. In addition, different exposure and lighting change across the frames can lead to flickering artifacts in the results. Recent works show that using a high-dimensional embedding~\cite{tancik2020fourier}, or using periodic activation functions~\cite{sitzmann2020implicit} can help recover fine details. However, we found that the above method is causing over-fitting in the training views and a lack of accuracy in the test views on our NeuLF representation. Learning how to recover the fine details of the 2PP light field representation can be an interesting direction in the future.

\begin{figure}[!h]
    \centering
    \includegraphics[width=\linewidth]{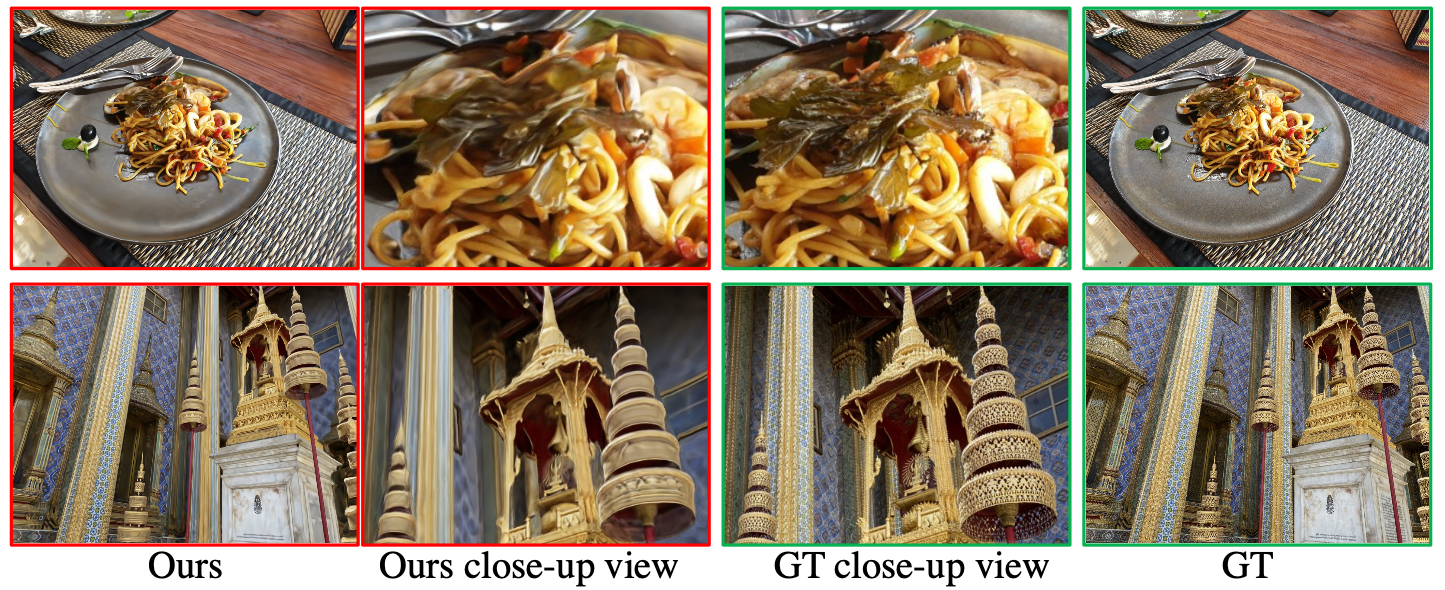}
    \caption{Our failure cases. We show synthesized novel views and groundtruth. On the close-up views, we can see that our results lack fine details compared with groundtruth.}
    \label{fig:fail}
\end{figure}

\section{Conclusion}
We propose a novel view synthesis framework called Neural 4D Light Field (NeuLF). Unlike NeRF, we represent a continuous scene using a 4D light field and train an MLP network to learn this mapping from input posed images. By limiting novel view synthesis to include only front views, NeuLF can achieve a comparable quality level as NeRF, but achieves a 1000x speedup. Moreover, because the speedup is enabled by modeling the color of light rays, NeuLF does not need additional storage for acceleration. To optionally output per-ray depth, we propose two loss terms: multi-view consistency loss and depth loss. This enables synthetic auto-refocus effect. We demonstrate state-of-the-art novel view synthesis results, especially for scenes with reflection and refraction. We also provide a study to show the effectiveness of our method with much fewer input images compared with NeRF and NeX.


\section{Limitations and Future Work}
There are several limitations to our approach. First, the novel viewpoints are limited to be on the one side of the two light slabs. In the future, we would like to extend the method to use more flexible 4D parameterizations such as multiple two planes, two cylindrical surfaces, or two spherical surfaces. By assuming the color is constant along a ray in free space, NeuLF cannot model rays that are blocked by the scene itself; therefore, novel viewpoints are always outside of the convex hull of the scene. This is an inherited limitation from light field.

Instead of using a 4D parameterization, lower-dimensional parameterization for specific applications can also be used. For example, in the work of concentric mosaic~\cite{shum1999rendering}, by constraining camera motion to planar concentric circles, all input image rays are indexed in three parameters. By adopting this parameterization, a more compact representation of the scene can be achieved, which potentially runs even faster than a 4D parameterization.


{Free-viewpoint video can be a straightforward extension of NeuLF from static scenes to dynamic ones. In the future, we would like to explore the possibility of including time in the formulation following \cite{li2021neural}}.

Although our simplified NeLF model can significantly improve the rendering speed compared with NeRF, it also has the limitations when it comes to 3D scene structure recovery. In the future, we would like to extend our work to reconstruct the surface by using existing approaches such as Shape from Light Field (SfLF) techniques~\cite{heber2017neural}. 



{\small
\bibliographystyle{ieee_fullname}
\bibliography{egbib}
}

\end{document}